\documentclass[10pt,twocolumn,letterpaper]{article}

\usepackage{cvpr}              

\usepackage{multirow}
\usepackage{graphicx}
\usepackage{color}
\usepackage{colortbl}
\usepackage{multirow}
\usepackage{mathtools}
\usepackage[table,xcdraw]{xcolor}
\definecolor{cvprblue}{rgb}{0.21,0.49,0.74}
\usepackage{siunitx}
\usepackage{adjustbox}
\usepackage{algorithm}
\usepackage{algpseudocode}
\usepackage{pythonhighlight}
\usepackage[symbol]{footmisc}

\usepackage{float}
\usepackage{multicol}
\usepackage{dblfloatfix}
\usepackage{bm}

\newcommand{\Skip}[1]{}

\usepackage[pagebackref,breaklinks,colorlinks,citecolor=cvprblue]{hyperref}

\def\paperID{11690} 
\def\confName{CVPR}
\def\confYear{2024}

\title{SemCity: Semantic Scene Generation with Triplane Diffusion
}

\author{Jumin Lee$^{1}$\footnotemark[1]\,\,\,\,\,
Sebin Lee$^{1}$\footnotemark[1]\,\,\,\,\,
Changho Jo$^2$\,\,\,\,\,
Woobin Im$^1$\,\,\,\,\,
Juhyeong Seon$^1$\,\,\,\,\,
Sung-Eui Yoon$^1$\\\\
$^1$KAIST\,\,\,\,\,\,\,\,\,\,\,\,\,\,\,\,\,\,$^2$Neosapience, Inc.
}

\begin{document}


\twocolumn[{%
\renewcommand\twocolumn[1][]{#1}%
\maketitle
\begin{center}
\vspace{-5mm}
\centerline{\includegraphics[trim={0cm 9.98cm 0.55cm 0cm},clip,width=1.0\linewidth]{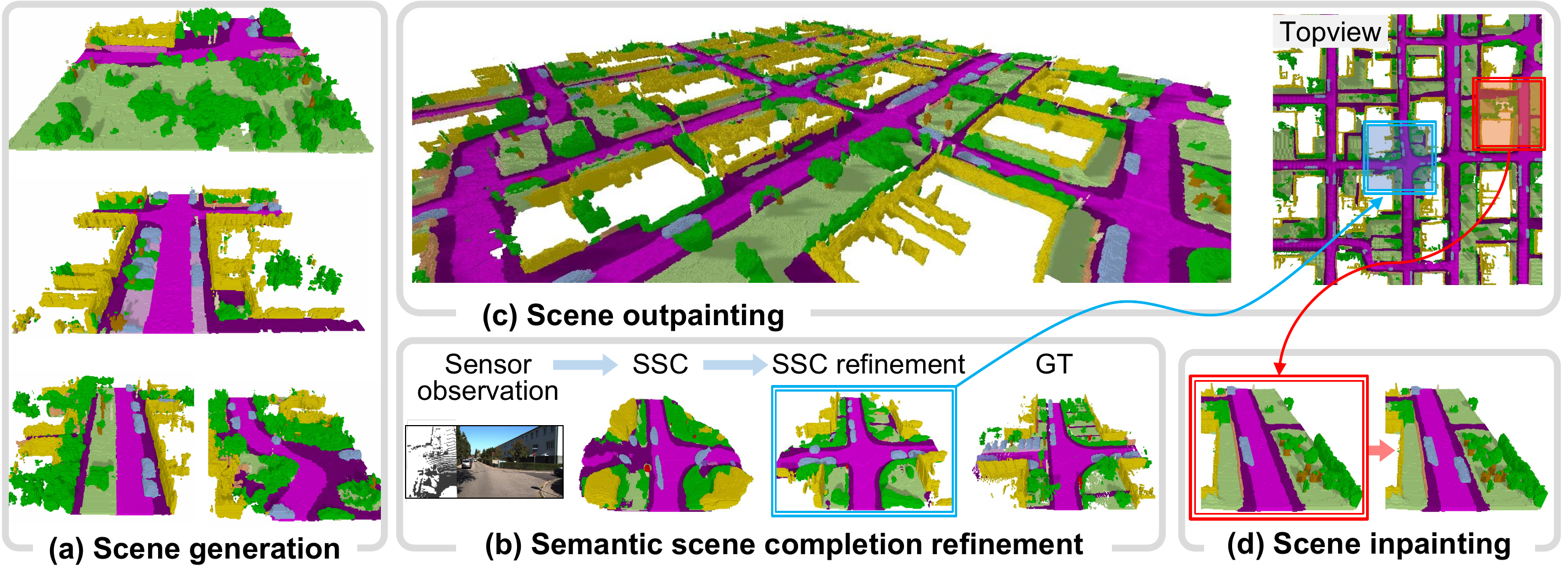}}
   \captionof{figure}{
   We introduce a diffusion framework, \textit{SemCity}, designed for generating semantic scenes in real-world outdoor environments as shown in  $\text{(a)}$.
   We extend our diffusion model to various practical tasks: semantic scene completion refinement, scene outpainting, and scene inpainting.
   For instance, the comprehensive scenario is displayed in $\text{(b)} \! \rightarrow \! \text{(c)} \! \rightarrow \! \text{(d)}$:
   the refined scene (SSC refinement) $\text{(b)}$ is outpainted to a broader scene $\text{(c)}$; then, an object (in this case, a car) is seamlessly integrated into the scene via our inpainting process $\text{(d)}$.
   }
    \label{fig:intro}
\end{center}%
}]


\footnotetext[1]{Both authors contributed equally to this work as co-first authors.}

\begin{abstract}
   We present ``SemCity," a 3D diffusion model for semantic scene generation in real-world outdoor environments.
Most 3D diffusion models focus on generating a single object, synthetic indoor scenes, or synthetic outdoor scenes, 
while the generation of real-world outdoor scenes is rarely addressed.
In this paper, we concentrate on generating a real-outdoor scene through learning a diffusion model on a real-world outdoor dataset.
In contrast to synthetic data, real-outdoor datasets often contain more empty spaces due to sensor limitations, causing challenges in learning real-outdoor distributions.
To address this issue, we exploit a triplane representation as a proxy form of scene distributions to be learned by our diffusion model.
Furthermore, we propose a triplane manipulation that integrates seamlessly with our triplane diffusion model. 
The manipulation improves our diffusion model's applicability in a variety of downstream tasks related to outdoor scene generation such as scene inpainting, scene outpainting, and semantic scene completion refinements.
In experimental results, we demonstrate that our triplane diffusion model shows meaningful generation results compared with existing work in a real-outdoor dataset, SemanticKITTI.
We also show our triplane manipulation facilitates seamlessly adding, removing, or modifying objects within a scene.
Further, it also enables the expansion of scenes toward a city-level scale.
Finally, we evaluate our method on semantic scene completion refinements where our diffusion model enhances predictions of semantic scene completion networks by learning scene distribution.
Our code is available at \url{https://github.com/zoomin-lee/SemCity}. 

\end{abstract}


\section{Introduction}
Diffusion models~\cite{ho2020denoising} have emerged as a promising generation tool owing to its state-of-the-art generation results in image domain~\cite{saharia2022palette, saharia2022photorealistic}.
This advance has led to active exploration in extending diffusion models to 3D data generation across both academic and industrial groups.
In the 3D domain, diffusion models have shown remarkable capabilities in generating diverse 3D forms (\textit{e.g.}, voxels, meshes)~\cite{Li_2023_CVPR, liu2023meshdiffusion}.
While those 3D diffusion models primarily aim to craft a single object, generating scenes consisting of multiple objects remains a relatively unexplored area in the 3D diffusion domain.

Scene generative diffusion models focus on crafting both geometrically and semantically coherent environments.
Compared with a single object generation, generating a scene with multiple objects requires an understanding of more complex geometric and semantic structure due to a larger spatial extent~\cite{tang2023diffuscene}.
There are primarily two streams of scene generative diffusion models, each tailored to either indoor or outdoor settings.
In particular, the outdoor environments 
have inherent challenges caused by a broader landscape than indoor ones.

We propose to leverage a triplane representation~\cite{chan2022efficient} for broader outdoor scenes, a method of factorizing 3D data onto three orthogonal 2D planes, 
as utilized in 3D object reconstruction and NeRF models~\cite{Wang_2023_CVPR, anciukevivcius2023renderdiffusion, mildenhall2021nerf}.
We excavate its advantages in addressing the data sparsity problem typically found in outdoor datasets due to the sensor limitations (\textit{e.g.}, occlusions, range constraints) in capturing outdoor scenes.
Triplane representation helps to reduce the inclusion of unnecessary empty information through the factorization of 3D data to 2D planes~\cite{chan2022efficient}. 
This efficiency in capturing relevant spatial detail makes it an effective tool for representing the many objects typically found in outdoor environments. 

In this paper, we design our diffusion framework based on triplane representations.
Our triplane autoencoder learns to compress a voxelized scene into a triplane representation by reconstructing semantic labels of the scene.
Following this, the triplane diffusion model is trained and used to generate new scenes, as shown in Fig.~\ref{fig:intro}(a), by creating novel triplanes based on the efficient representation.
Further, we propose a triplane manipulation method, which extends our triplane diffusion model toward several practical tasks (\textit{i.e.}, scene inpainting, scene outpainting, and semantic scene completion refinements) as shown in Fig.~\ref{fig:intro}(b-d).
Our method can seamlessly add, remove, and modify objects in real-outdoor scenes while maintaining the semantic coherence of environments.

Our contributions are summarized as follows:
\begin{itemize}
\item We disclose the applicability of the triplane representation through generating semantic scenes for real-outdoor environments and extend its views in practical downstream tasks: scene inpainting, scene outpainting, and semantic scene completion refinement.
\item We propose to manipulate triplane features during our diffusion process, facilitating seamlessly extending our method toward the downstream tasks.
\item We demonstrate that the proposed method significantly enhances the quality of generated scenes in real-world outdoor environments.
\end{itemize}

\section{Related Work}
\noindent \textbf{Diffusion Models.}
Diffusion models~\cite{ho2020denoising} learn data distributions via iterative denoising processes based on score functions~\cite{song2020score}.
Its generated results have shown remarkably realistic appearances with high fidelity and diversity in a variety of 2D image synthesis such as outpainting~\cite{saharia2022palette, zhang2023diffcollage}, inpainting~\cite{saharia2022palette, lugmayr2022repaint} and text-to-image generation~\cite{saharia2022photorealistic, ramesh2022hierarchical}.
Built upon these achievements, diffusion models have also been extended into the 3D domain, generating impressive results in various 3D shapes, including 
voxel grids~\cite{zhou20213d, Li_2023_CVPR}, 
point clouds~\cite{Luo_2021_CVPR, zeng2022lion, Yu_2023_ICCV},
meshes~\cite{liu2023meshdiffusion}, 
and implicit functions~\cite{jun2023shap, wu2023sin3dm, shue2023nfd, Shim_2023_CVPR}.
While these models can craft a single 3D object, our model focuses on generating a 3D scene composed of multiple objects using a categorical voxel data structure, which is a relatively under-explored area in the 3D diffusion domain.



\begin{figure*}[t]
\begin{center}
\centerline{\includegraphics[trim={0cm 12.72cm 0.0cm 0cm},clip,width=1.0\textwidth]{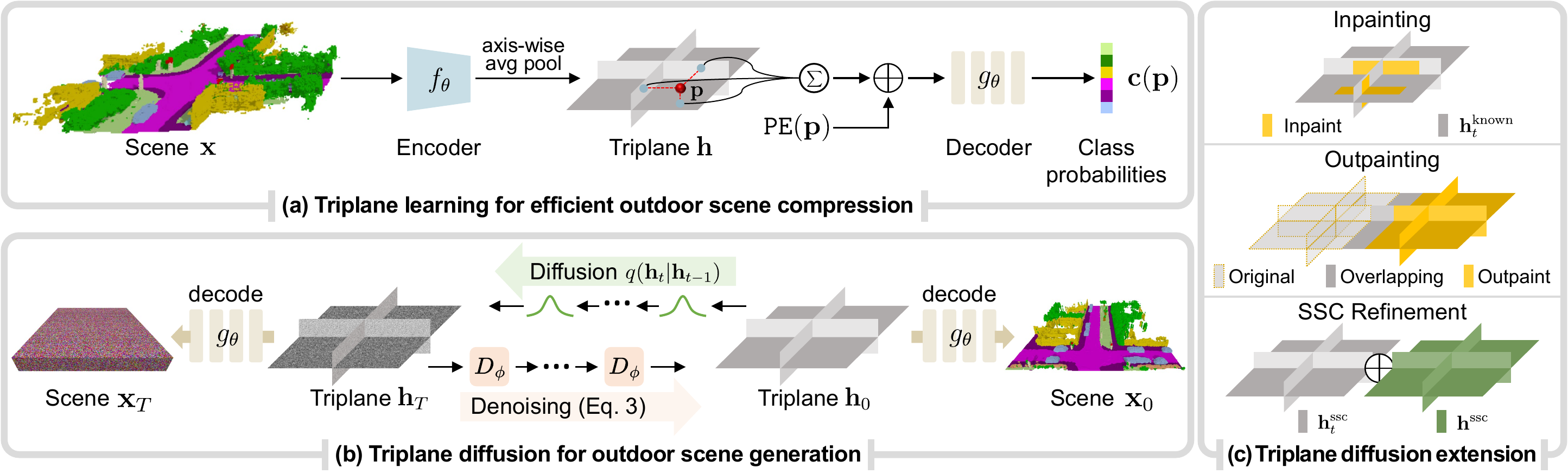}}
    \captionof{figure}{
    \textbf{Overview of ours.}
    (a) A 3D semantic map $\mathbf{x}$ is encoded by a triplane encoder $f_\theta$ and factorized to a triplane $\mathbf{h}$. The triplane coupled with a positional encoding $\texttt{PE}(\mathbf{p})$ is decoded by an implicit decoder $g_\theta$, resulting in class probabilities for each coordinate $\mathbf{p}$.
    (b) Our triplane diffusion model $D_\phi$ learns to generate a novel triplane for semantic scene generation via denoising diffusion process. 
    (c) We further extend our triplane diffusion beyond a simple scene generation toward various practical scenarios by manipulating triplanes in (b).
    }\vspace{-5mm}
    \label{fig:method}
\end{center}
\end{figure*} 
\paragraph{Diffusion Models for Scene Generation.}
In contrast to a single object generation, 
scene generation involves an understanding of the larger 3D space, 
causing more semantic and geometric complexities~\cite{tang2023diffuscene}.
Diffusion models for scene generation have been studied in both indoor and outdoor environments. 
In indoor settings, diffusion models aim to learn distributions of relations among objects by representing them as scene graphs~\cite{johnson2018image}.
The scene graphs contain object attributes (\textit{e.g.}, location, orientation, and size), capturing the intricate inter-object relationships within bounded spaces~\cite{tang2023diffuscene, zhai2023commonscenes}. 
For outdoor scenes, the challenges are distinct, frequently including a lot of empty areas (\textit{e.g.}, sky, open areas) resulting from the broader landscapes. 
Traditional approach~\cite{lee2023} has relied on discrete diffusion methods~\cite{hoogeboom2021argmax} on voxel space, necessitating a detailed representation of every air volume.

In this paper, we demonstrate that triplane diffusion is highly effective for generating real-outdoor scenes. 
By abstracting 3D spaces into three orthogonal 2D planes, triplane representation~\cite{chan2022efficient} effectively captures the vastness of outdoor environments, predominantly composed of air.
Beyond its data efficiency, the triplane excels in focusing on other significant objects (\textit{e.g.}, vehicles, buildings) by allocating lesser attention to less informative elements like air. 
Our approach stands in stark contrast to prior work~\cite{lee2023}, which were constrained to synthetic datasets with considerably less empty space.
In real datasets, inherent sensor limitations, such as a limited field-of-view, a limited ranges, and the inability to capture occluded areas like the rears of buildings, lead to a prevalence of empty space.
Furthermore, we emphasize the versatility of our framework by demonstrating its extension to various downstream tasks, including scene inpainting, outpainting, and semantic scene completion refinement.

\paragraph{3D Inpainting and Outpainting.} 
In 3D inpainting, the primary objective is to fill in missing portions or modify existing elements of 3D data while maintaining geometric consistency.
Most existing works concentrate on single-object inpainting~\cite{Anciukevicius_2023_CVPR, 9197608, lei2022generative};
for instance, they seamlessly transit a 3D chair's leg count from three to four.
Contrary to inpainting, 3D outpainting is to extrapolate a given scene over an unobserved space.
The existing work~\cite{abbasi2019deep} focuses on scene outpainting within bounded indoor environments such as rooms.
While in 2D images, inpainting and outpainting are not restricted to a single object~\cite{yu2018generative, jo2021n,cheng2022inout}.
Likewise, in 3D space, we focus on scene-level inpainting, which seamlessly adds, removes, or modifies objects in a scene.
Further, our scene-level outpainting is not constrained to the bounded scenes; 
we extrapolate the outdoor scene from sensor range (\textit{e.g.}, LiDAR) to city-scale.

\paragraph{Semantic Scene Completion.}
Semantic scene completion (SSC)~\cite{song2017semantic} is pivotal for 3D scene understanding, where it jointly infers completion and semantic segmentation of the 3D scene
from sensor observations such as RGB images~\cite{cao2022monoscene, miao2023occdepth, li2023voxformer} or point clouds~\cite{yang2021semantic, xia2023scpnet, yan2021sparse}.
In addition, SSC plays a crucial role in supporting comprehensive autonomous navigation systems, notably in essential downstream tasks like path planning~\cite{9341347, wang2019efficient, chaplot2020object} and map construction~\cite{cui20203d, wang2019autonomous}.
Despite significant progress in the field, a persistent challenge is the semantic and geometric discrepancies between the SSC-estimated scenes and their real counterparts, as illustrated in Fig.~\ref{fig:intro}(c). 
These discrepancies can undermine the performance of downstream tasks.
Our triplane diffusion model can help to bridge this gap by exploiting 3D scene priors.
This approach enhances the reliability and effectiveness of SSC, which is expected to improve its application in autonomous navigation systems.

\section{Method}
In this section, we elucidate our triplane diffusion model and its extensions.
Our triplane diffusion model aims to synthesize novel real-outdoor scenes by generating a triplane, a proxy representation, which effectively addresses the inherent challenges of real-outdoor scene synthesis.
This triplane representation is learned by our triplane autoencoder, which abstracts
the geometric and semantic intricacies of a scene into three orthogonal 2D feature planes, namely, the $xy$, $xz$, and $yz$ planes (Sec.~\ref{ae}).
Then, our diffusion model learns triplane distributions of scenes, generating novel triplanes (Sec.~\ref{diffusion}).
We extend our triplane diffusion model toward various practical scenarios: 
scene inpainting, scene outpainting, and semantic scene completion refinements (Sec.~\ref{refine}).

\subsection{Representing a Semantic Scene with Triplane}
\label{ae}
To represent a 3D scene as a triplane, our triplane autoencoder learns 
to compress a 3D scene into a triplane representation as shown in Fig.~\ref{fig:method}(a). 
The autoencoder consists of two modules: 
(1) an encoder $f_\theta$ yielding a triplane, 
and (2) an implicit multi-layer perceptron (MLP) decoder $g_\theta$ for reconstruction from the triplane.

The encoder $f_\theta$ takes a voxelized scene $\mathbf{x} \! \in \! \mathbb{R}^{X \times Y \times Z}$ containing $N$ classes
within a spatial grid of resolution $X \! \times \! Y \! \times \! Z$.
It then yields an axis-aligned triplane representation $\mathbf{h} \! = \! [\mathbf{h}^{xy}, \mathbf{h}^{xz}, \mathbf{h}^{yz} ]$.
The triplane consists of three planes, each characterized by distinct dimensional properties:
$\mathbf{h}^{xy} \! \in \! \mathbb{R}^{C_h \! \times \! X_h \! \times \! Y_h}$, $\mathbf{h}^{xz} \! \in \! \mathbb{R}^{C_h \! \times \! X_h \! \times \! Z_h}$, and $\mathbf{h}^{yz} \! \in \! \mathbb{R}^{C_h \! \times \! Y_h \! \times \! Z_h}$,
where $C_h$ stands for a feature dimension, and $X_h$, $Y_h$, and $Z_h$ denote spatial dimension of the triplane.
During the encoding phase, a 3D feature volume is extracted by 3D convolutional layers from the scene $\mathbf{x}$, resulting in the triplane via axis-wise average pooling.
Given a 3D coordinate $\mathbf{p} \! = \! (x, y, z)$, 
the triplane is interpreted as a summation of vectors bilinearly interpolated from each plane: 
$\mathbf{h}(\mathbf{p}) \! = \! \mathbf{h}^{xy}(x, y) \! + \! \mathbf{h}^{xz}(x, z) \! + \! \mathbf{h}^{yz}(y, z)$.

To reconstruct the 3D scene $\mathbf{x}$, 
we decode the encoded triplane $\mathbf{h}$ 
with an implicit MLP decoder $g_\theta$ that predicts semantic class probabilities.
The decoder takes the triplane vector $\mathbf{h}(\mathbf{p})$ 
with its sinusoidal positional embedding $\texttt{PE}(\mathbf{p})$~\cite{mildenhall2020nerf},
resulting in class probabilities $\mathbf{c}(\mathbf{p}) \! = \! g_\theta(\mathbf{h}(\mathbf{p}), \texttt{PE}(\mathbf{p})) \! \in \! [0, 1]^{N} $.
The positional embedding produces high-frequency features according to the coordinates $\mathbf{p}$, 
which helps the implicit decoder $g_\theta$ represent high-frequency scene contents~\cite{wang2023pet}.

The encoder $f_\theta$ and the MLP decoder $g_\theta$ are trained with the autoencoder loss $\mathcal{L}_\text{AE}$ and scene label $\mathbf{x}(\mathbf{p})$ as:
\begin{equation}\label{eq:ae_loss}
    \mathcal{L}_\text{AE} 
    = \mathbb{E}_{\mathbf{p}\sim \mathcal{P} } [  
        \ell_\text{CE}(\mathbf{c}(\mathbf{p}), \mathbf{x}(\mathbf{p}))  
        + \lambda \hspace{0.10em} \ell_\text{LZ}(\mathbf{c}(\mathbf{p}), \mathbf{x}(\mathbf{p})) 
    ],
\end{equation}
where $\lambda$ is a loss weight, and $\mathcal{P}$ is the set of grid coordinates of the scene. 
We use the weighted cross-entropy loss $\ell_\text{CE}$~\cite{roldao2020lmscnet} 
and the Lov\'{a}sz-softmax loss $\ell_\text{LZ}$~\cite{berman2018lovasz}
to learn imbalanced semantic distributions of the scene.

\subsection{Triplane Diffusion}
\label{diffusion}
Based on the triplane representation of the 3D semantic scene,
our triplane diffusion model $D_\phi$ learns to generate a novel triplane 
through denoising diffusion probabilistic models~\cite{ho2020denoising} as shown in Fig.~\ref{fig:method}(b).
This triplane generation leads to generation of 3D scene
through decoding the generated triplane with the implicit MLP decoder $g_\theta$.
Through the $\textit{\textbf{x}}_0$-parameterization~\cite{austin2021structured},
the diffusion model $D_\phi$ is trained to reconstruct the triplane $\mathbf{h}$
given its corrupted triplane $\mathbf{h}_t$ sampled from a diffusion process 
$q(\mathbf{h}_t|\mathbf{h}) \! = \! \mathcal{N}(\sqrt{\bar{\alpha}_t} \, \mathbf{h}, (1 - \bar{\alpha}_t) \, \mathbf{I})$,
where $\mathcal{N}$ is the Gaussian distribution, $\bar{\alpha}_t = \prod_{i=1}^{t} \alpha_i$, and $\alpha_t=1-\beta_t$ with a variance schedule $\beta_t$.
The diffusion process $q(\mathbf{h}_t|\mathbf{h})$ is derived from the Markovian chain rule with a single step's diffusion process $q(\mathbf{h}_t | \mathbf{h}_{t-1}) \! = \! \mathcal{N}(\sqrt{1-\beta_t} \mathbf{h}_{t-1}, \beta_t \mathbf{I})$.
Thus, the triplane diffusion loss is defined as:
\begin{equation}
\label{eq:diffusion_loss}
    \mathcal{L}_\text{D} = \mathbb{E}_{t \sim \mathcal{U}(1,T)} ||\mathbf{h} - D_{\phi}(\mathbf{h}_{t}, t)||_{p},
\end{equation}
where $T$ is the number of denoising steps, and $p$ represents the order of the norm. The timestep $t$ is sampled from the discrete uniform distribution $\mathcal{U}$.

After training, 
the diffusion model $D_\phi$ generates a novel triplane $\mathbf{h}_0$
via the iterative DDPM generation process~\cite{ho2020denoising} starting from $\mathbf{h}_T \sim \mathcal{N}(\mathbf{0}, \mathbf{I})$:
\begin{equation}
\label{eq:generation}
    \mathbf{h}_{t-1}  \sim 
    \mathcal{N} \! \left(
        \gamma_t \mathbf{h}_t 
        + \delta_t D_\phi(\mathbf{h}_t, t),
        \beta_t^2 \hspace{0.10em} \mathbf{I}
    \right),
\end{equation}
with $\gamma_t \! \coloneqq \! \sqrt{\alpha_t}(1 - \bar{\alpha}_{t-1}) / ( 1 - \bar{\alpha}_t )$
and $\delta_t \! \coloneqq \! \sqrt{\bar{\alpha}_{t-1}}\beta_t / ( 1 - \bar{\alpha}_t )$.
From the generated triplane $\mathbf{h}_0$,
we generate a novel 3D semantic scene $\mathbf{x}_0$ 
by querying coordinates $\mathbf{p}$ 
to the implicit decoder, \textit{i.e.}, $g_\theta(\mathbf{h}_0(\mathbf{p}), \texttt{PE}(\mathbf{p}))$.

\subsection{Applications with Triplane Manipulation}
Building on the triplane diffusion process (Sec.~\ref{diffusion}),
we propose a triplane manipulation that allows our model to facilitate a variety of practical downstream tasks with few modifications, as illustrated in Fig.~\ref{fig:method}(c).

\paragraph{Scene Inpainting.}
\label{sec3:inpainting}
Our scene inpainting randomly edits a 3D scene, 
seamlessly adding, modifying, or removing objects 
while maintaining the consistency and realism of the scene. 
For instance, the inpainting includes scenarios where cars or sidewalks appear and then disappear, or vice versa, as shown in Fig.~\ref{fig:intro}(d). 
Inspired by the RePaint sampling strategy~\cite{lugmayr2022repaint},
we propose a 3D-aware inpainting approach
with semantic coherence.
RePaint focuses on the image domain without explicitly considering the fidelity of the underlying 3D scene.
To facilitate 3D-aware inpainting, 
we inpaint the triplane, serving as a compact proxy representation for the scene.
We define a binary spatial \textit{trimask} $\mathbf{m} \! = \! [ \mathbf{m}^{xy}, \mathbf{m}^{xz}, \mathbf{m}^{yz} ]$ covering inpainting regions on triplane space, allowing us to control the generation process on the masked region.
The trimask $\mathbf{m}$ is set to have ones for inpainting regions and zeros for others.
We override the $t$-th triplane $\mathbf{h}_{t} \! = \! [ \mathbf{h}_t^{xy}, \mathbf{h}_t^{xz}, \mathbf{h}_t^{yz} ]$ of the generation process (Eq.~\ref{eq:generation}) as follows:
\begin{equation}
\label{eq:inpainting}
    \mathbf{h}_{t} \leftarrow 
    \mathbf{m} \otimes \mathbf{h}_{t}
    + (\mathbf{1} - \mathbf{m}) \otimes \mathbf{h}_{t}^\text{known},
\end{equation}
where $\otimes$ is the element-wise product, 
and the known triplane $\mathbf{h}_t^{\text{known}}$ for intact regions is sampled from the diffusion process,
\textit{i.e.}, $q(\mathbf{h}_t^\text{known} | \mathbf{h})  \! \coloneqq \! q(\mathbf{h}_t | \mathbf{h})$, 
which adheres to a known Gaussian distribution.

\paragraph{Scene Outpainting.}
\label{outpaint}
Our scene outpainting extends the boundaries of the 3D scene without additional training as with inpainting.
To seamlessly outpaint a scene, the regions to be extended should be conditioned on the original scene. 
We propose to inject this intuition into triplane representation as shown in Fig.~\ref{fig:method}(c).
For covering regions to be outpainted, our diffusion model generates a novel triplane that is partially overlapped with the original triplane.
We implement this with the concept of the trimask $\mathbf{m}$ and the known triplane $\mathbf{h}_t^\text{known}$ as in our scene inpainting.
The trimask $\mathbf{m}$ covers regions to be outpainted, and the $\mathbf{h}_t^\text{known}$ is obtained from the intersections of triplanes between original and outpainted regions.
Based on the outpainting strategy, we extend a given scene toward cardinal and intercardinal directions, facilitating the creation of an unbounded scene.
While our triplane diffusion model is trained using triplanes of a fixed size, 
it demonstrates the capability to outpaint scenes to be several times bigger than the original scene, as illustrated in Fig.~\ref{fig:intro}(c).
\begin{figure*}[t]
\begin{center}
     \includegraphics[trim={0cm 0cm 0cm 0cm},clip,width=1.0\linewidth]{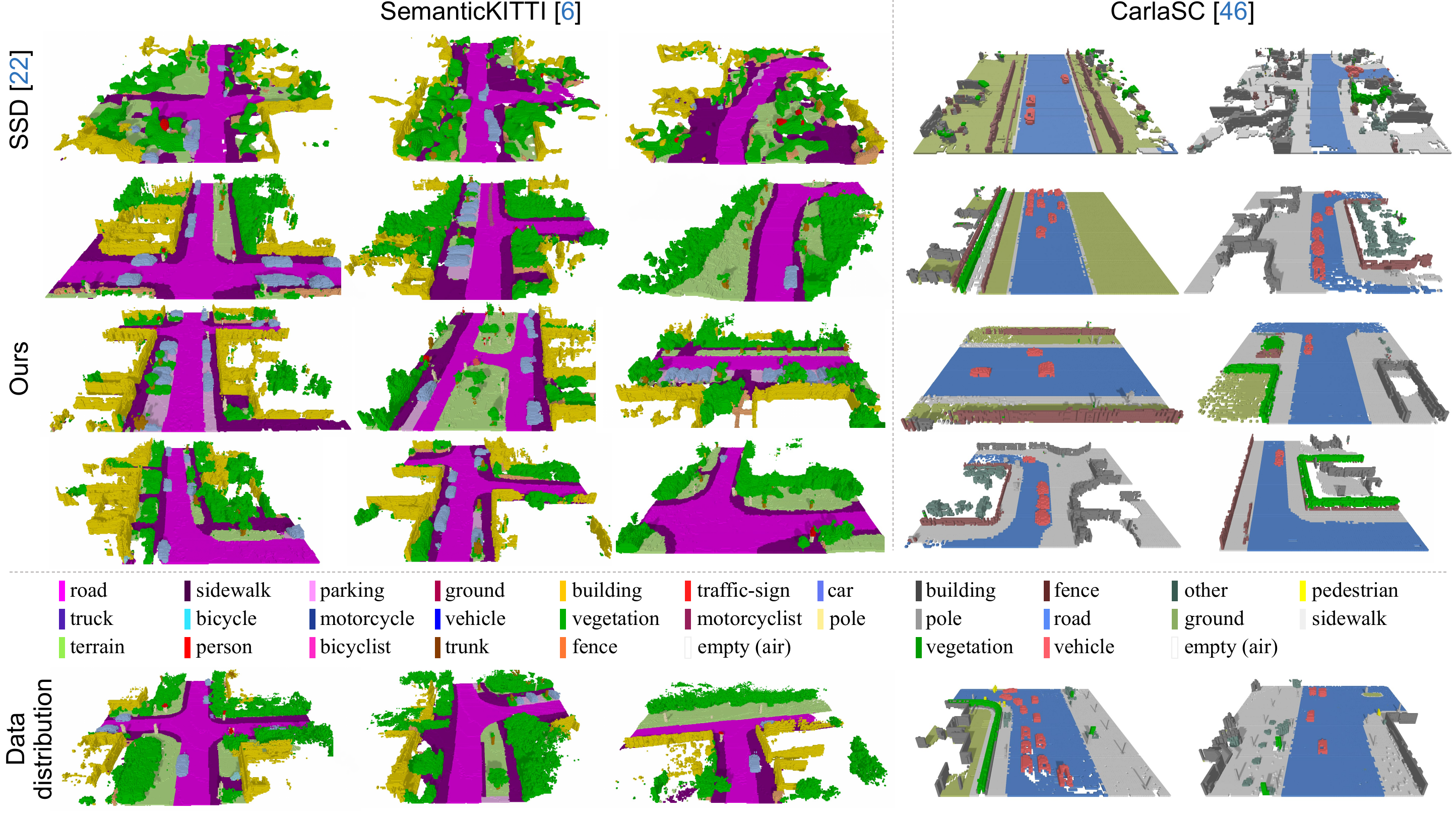}
    \vspace{-5mm}
    \captionof{figure}{
        \textbf{Scene generation results} using both real and synthetic outdoor datasets -- SemanticKITTI~\cite{semantickitti} and CarlaSC~\cite{carlasc}.
        Our results showcase the effective generation of overall structures, including roads and buildings, along with detailed objects such as cars.
    }
    \label{fig:gen}
\end{center} \vspace{-5mm}
\end{figure*}

\paragraph{Semantic Scene Completion Refinement.} 
\label{refine} 
SSC models complete and segment a 3D scene from sensor observations such as images or point clouds.
We observe that the SSC results show a geometric and semantic discrepancy compared with data distributions, as shown in Fig.~\ref{fig:intro}(b).
We extend our triplane diffusion model to refine predictions of SSC models toward reducing the discrepancy.
To effectively condition our triplane diffusion scheme to SSC model's prediction $\mathbf{x}^\text{ssc}$, 
we utilize its triplane representation $\mathbf{h}^\text{ssc} \! = \! f_\theta(\mathbf{x}^\text{ssc})$ derived by our triplane encoder $f_\theta$.
We extend our triplane diffusion scheme with a simple modification of the triplane $\mathbf{h}_t$ in the diffusion loss (Eq.~\ref{eq:diffusion_loss}) and the generation process (Eq.~\ref{eq:generation}) as follows:
\begin{equation}
\label{eq:ssc_refinement}
    \mathbf{h}_{t} = 
    \mathbf{h}_{t}^\text{ssc} 
    \oplus \mathbf{h}^\text{ssc},
\end{equation}
where $\oplus$ is concatenation, and $\mathbf{h}^\text{ssc}_t$ is a $t$-th diffused triplane sampled by the DDPS diffusion process~\cite{lai2023denoising} with the SSC prediction's triplane $\mathbf{h}^\text{ssc}$.

\section{Experiments}\label{sec:expr}
\subsection{Experimental Details}
\paragraph{Training Dataset.}
\label{dataset}
We validate our method on the SemanticKITTI~\cite{semantickitti} and CarlaSC~\cite{carlasc} datasets.
SemanticKITTI provides 3D semantic scenes of real-outdoor environments with labels for 20 semantic classes.
Each scene is represented by a voxel grid of $256\times 256\times 32$, covering an area of \SI{51.2}{\meter} in front of the car, extending \SI{51.2}{\meter} on each side, and reaching up to a height of \SI{6.4}{\meter}.
The dataset retains object motion traces as a result of sensor frame integration, which is employed to establish a dense ground truth.
In contrast, CarlaSC is a synthetic dataset that provides 3D semantic outdoor scenes without the trace of moving objects.
The dataset contains annotated 11 semantic classes with a voxel grid of $128\times128\times8$ and covers a distance of \SI{25.6}{\meter} in front and behind the car, \SI{25.6}{\meter} laterally on each side, and \SI{3}{\meter} in height.

\paragraph{Implementation Details.}
Our experiments are deployed on a single NVIDIA RTX 3090 GPU with a batch size of 4 for the triplane autoencoder and 18 for the triplane diffusion model.
For the triplane autoencoder, the input scene is encoded to triplane with a spatial resolution $(X_h, Y_h, Z_h)=(128,128,32)$, and the feature dimension $C_h$ is 16.
The loss weight $\lambda$ in Eq.~\ref{eq:ae_loss} is set to 1.0.
The order of the norm $p$ in Eq.~\ref{eq:diffusion_loss} is set to 1 for SSC refinements and 2 for other cases.
For the diffusion model, the learning rate is initialized to 1e-4 and then decreases linearly. 
During the diffusion process, we use the default settings~\cite{rombach2022high} with 100 time steps ($T$).
For our triplane inpainting and outpainting, we employ the RePaint sampling strategy~\cite{lugmayr2022repaint} as a reference and perform a repaint with 5 resampling and a jump size of 20. 

\paragraph{Evaluation Metrics.}
Following the scene generation works~\cite{tang2023diffuscene, zhai2023commonscenes}, we evaluate the performance of semantic scene generation by examining both the diversity and fidelity of 3D semantic scenes within the rendered images.
We use recall to evaluate diversity, while precision and inception score (IS) are used to evaluate fidelity.
The Fréchet Inception Distance (FID)~\cite{heusel2017gans} and Kernel Inception Distance (KID)~\cite{binkowski2018demystifying} metrics are also utilized, as they reflect the combined effect of both diversity and fidelity on scene quality~\cite{dhariwal2021diffusion}.
In terms of semantic scene completion (SSC) refinement performance, we follow the protocols defined in SSC works~\cite{xia2023scpnet, cao2022monoscene, miao2023occdepth, yan2021sparse}.
The Intersection-over-Union (IoU) metric is used to quantify scene completeness, while the mean IoU (mIoU) provides a measure for the quality of semantic segmentation.
These metrics together enable a comprehensive evaluation of how well the semantic scene completion methods perform in terms of accurately filling in and labeling the scene components.

\begin{table}[]
\renewcommand{\tabcolsep}{1.81mm}
\small{
\begin{tabular}{clccccc}
\hline
\multicolumn{2}{l}{Model}   & FID $\downarrow$   & KID $\downarrow$ & IS $\uparrow$   & Prec $\uparrow$ & Rec $\uparrow$  \\ \hline
\multicolumn{7}{l}{SemanticKITTI~\cite{semantickitti}}   \\
           & SSD~\cite{lee2023}          & 112.82 & 0.12 & 2.23 & 0.01 & 0.08 \\
           & SemCity (Ours)                 & 56.55  & 0.04 & 3.25 & 0.39 & 0.32 \\ \hline
\multicolumn{7}{l}{CarlaSC~\cite{carlasc}}      \\
           & SSD~\cite{lee2023}         & 87.39 & 0.09 & 2.44 & 0.14 & 0.07 \\
           & SemCity (Ours)                 & 40.63  & 0.02  & 3.51 & 0.31 & 0.09 \\ \hline
\end{tabular}} 
\caption{
    \textbf{Results of semantic scene generation.} 
    Each metric is computed between the rendered image of the generated scene and the ones of the actual scene at a resolution of $1440 \times 2048$.
}\vspace{-1mm}
\label{tab:gen}
\end{table}

\subsection{Semantic Scene Generation}
Fig.~\ref{fig:gen} illustrates a qualitative comparison on the SemanticKITTI~\cite{semantickitti} and CarlaSC~\cite{carlasc} datasets.
SSD~\cite{lee2023} shows impressive results on the CarlaSC dataset.
However, its performance on the SemanticKITTI dataset, which is a real-world dataset, is notably constrained.
This limitation primarily arises from the SSD's voxel-based representation, which struggles with the more prevalent empty spaces in real-outdoor datasets compared to synthetic ones.
This issue is especially evident in the generation of buildings and roads, where SSD often fails to define boundaries accurately.
Furthermore, the model's limitations extend to representing finer structures, such as trunks and leaves, as well as traffic light poles and signals.
It also struggles with generating uniform shapes of vehicles, often resulting in irregular shapes.
In contrast, our method demonstrates the ability to effectively synthesize detailed scenes even on the real dataset, as illustrated in Fig.~\ref{fig:gen}.
It performs better than SSD~\cite{lee2023} in accurately capturing complex building shapes on the CarlaSC dataset. 
\begin{figure}[t]
 \centering
    \centering
    \captionsetup{type=figure}
        \includegraphics[trim={0cm 16.6cm 22.5cm 0cm},clip,width=1.0\columnwidth]{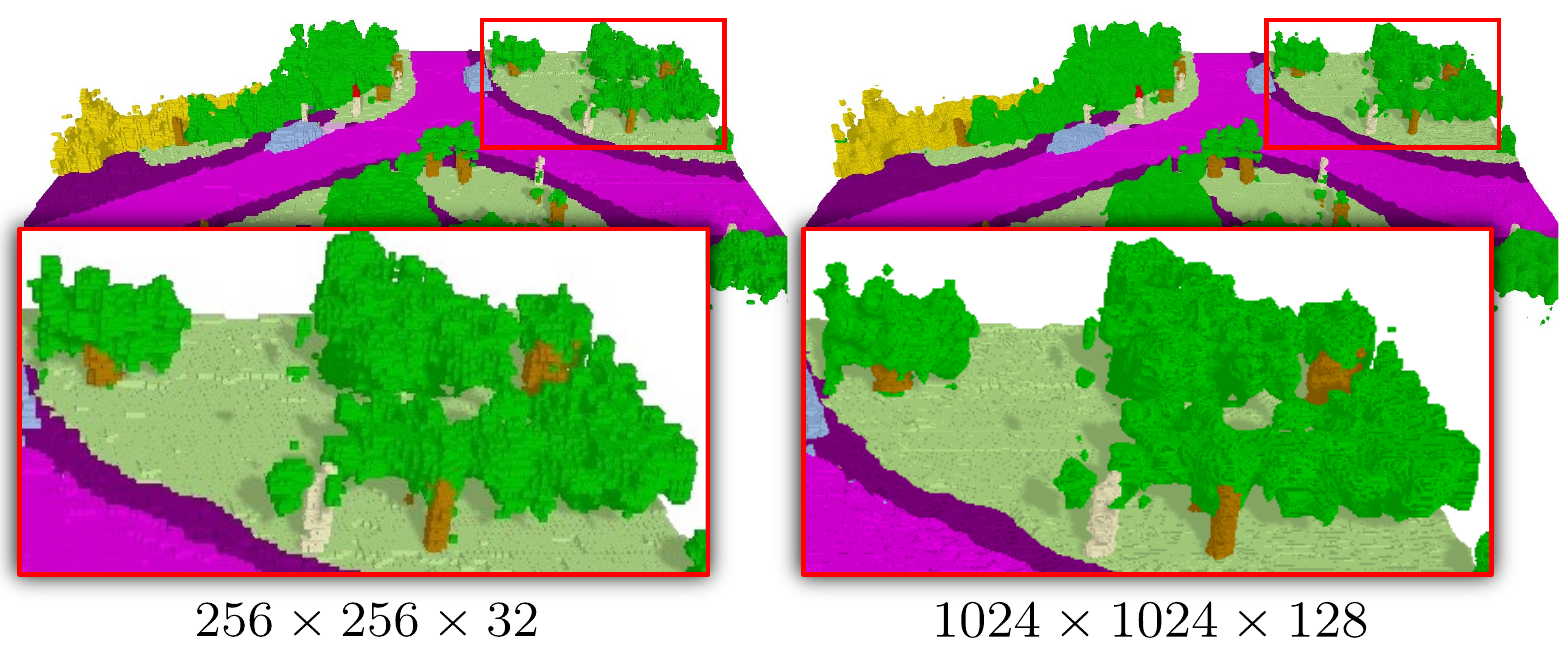}
        \caption{
            \textbf{Higher-resolution scene generation.}
            Building upon our implicit decoder, higher-resolution scene ($1024 \times 1024 \times 128$) can be generated compared with a resolution of training dataset ($256 \times 256 \times 32$).
        }\label{fig:implicit}  
\end{figure}
\begin{figure}[t]
 \centering
    \centering
    \captionsetup{type=figure}
        \includegraphics[trim={0cm 1.9cm 25.9cm 0cm},clip,width=1.0\columnwidth]{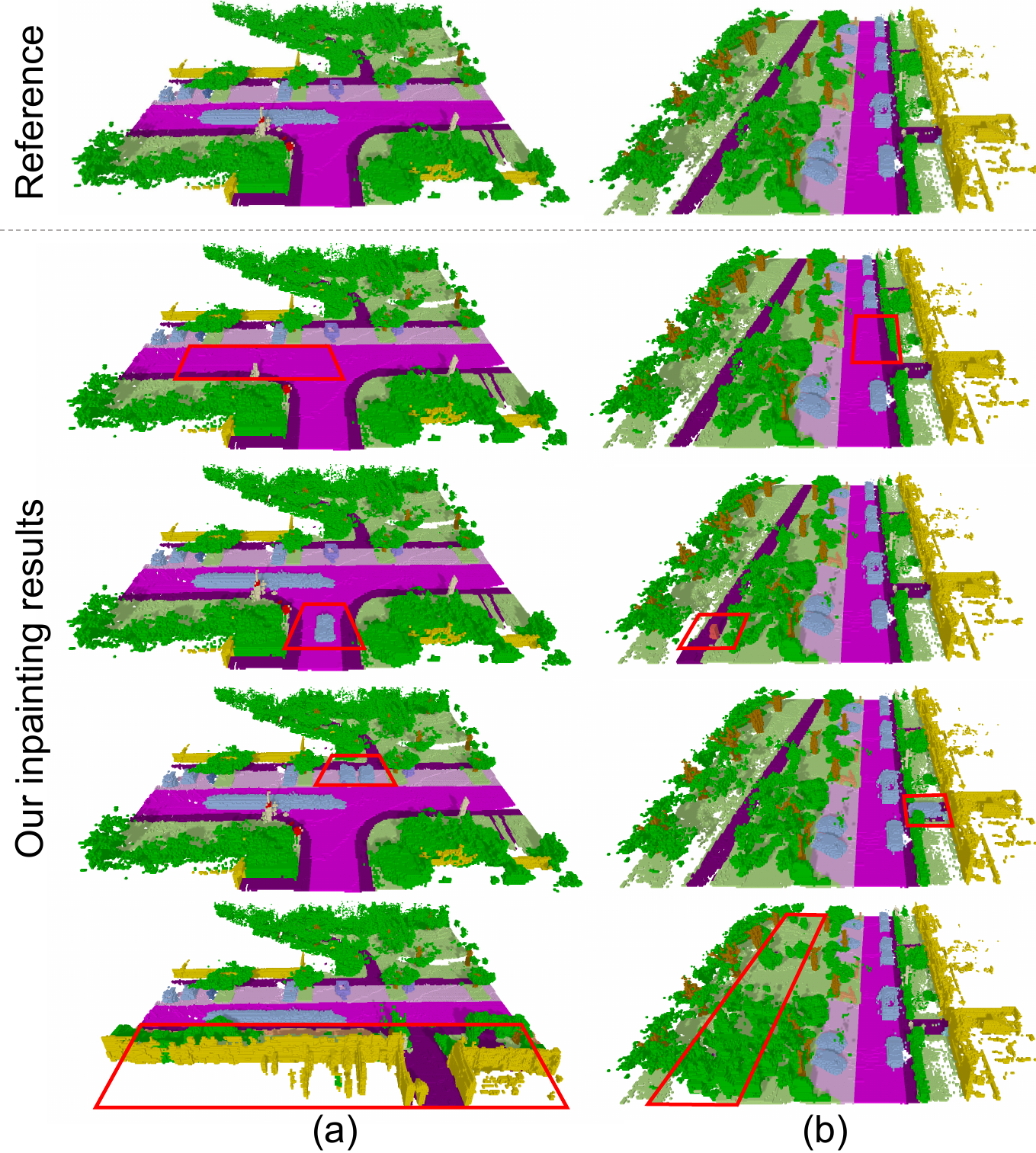}\vspace{-2mm}
        \caption{
            \textbf{Scene inpainting of our method.}
            The red boxes denote inpainting regions.
            (a) and (b) show our inpainting examples from reference images.
        }\label{fig:inpaint}  
\end{figure}
In addition, our method exhibits remarkable proficiency in generating the overall contours of roads and buildings, along with intricate details on the SemanticKITTI dataset.
Tab.~\ref{tab:gen} provides a detailed comparative evaluation using various metrics.
Our model shows significant improvements in both the fidelity and diversity of the generated scenes.
Moreover, our generated result is not tied to fixed resolution by means of the implicit neural representation, as depicted in Fig.~\ref{fig:implicit}.
For additional results, please refer to the Supplementary Material.
\begin{figure*}[t]
\begin{center}
\includegraphics[trim={0cm 10.2cm 0cm 0cm},clip,width=1.0\linewidth]{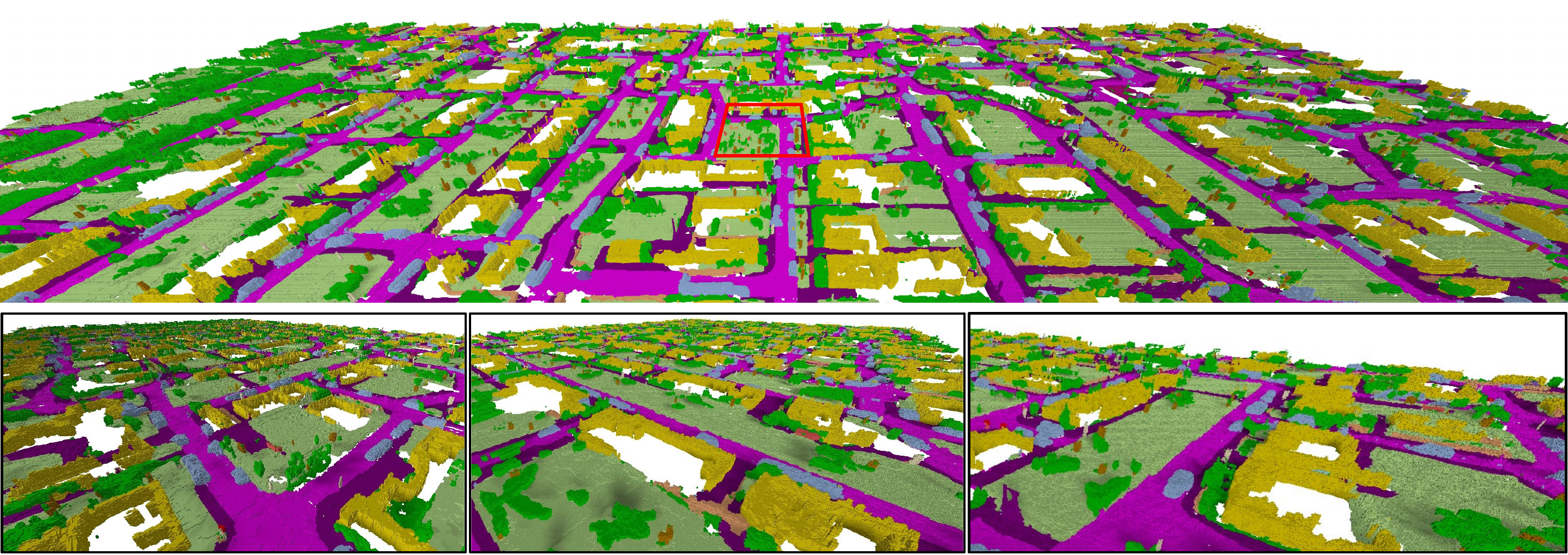}
\captionof{figure}{
    \textbf{Our scene outpainting results.} 
    The red box at the center of the figure in the first row represents the given scene for outpainting.
    The zoomed views of the outpainted scene are depicted in the second row.
    The outpainted scene is expanded from the given size of $256 \times 256 \times 32$ to $1792 \times 3328 \times 32$.
}\label{fig:city}\vspace{-2mm}
\end{center}%
\end{figure*}

\subsection{Applications of Triplane Diffusion}
\paragraph{Scene Inpainting.}
Fig.~\ref{fig:inpaint} presents the qualitative results of our inpainting, demonstrating its effectiveness in inpainting both small and large regions within a scene, while maintaining the coherence of 3D contexts.
In detail, the second row of \textbf{(a)} and \textbf{(b)} illustrates the model's seamless removal of vehicles, which harmonizes with the adjacent road.
The third row demonstrates the insertion of new entities — a vehicle in \textbf{(a)} and a person in \textbf{(b)} — that are contextually congruent with the reference scene.
The fourth row in \textbf{(a)} exemplifies the model's dual functionality in both modifying and adding vehicles within the scene.
The fifth row in both columns underscores the model's proficiency in modifying scenes.
Here, the model alters existing scene components, showcasing its ability to transform the overall ambiance of the scene.
These results show our model's adeptness not just in object-level inpainting but also in scene-level inpainting.
\begin{table}[]\renewcommand{\tabcolsep}{3.25mm}
\small{
\begin{center}
\begin{tabular}{llcc}
\hline
SSC Input                    & Method         & IoU $\uparrow$  & mIoU $\uparrow$ \\ \hline
\multirow{4}{*}{RGB}         & MonoScene~\cite{cao2022monoscene} & 37.12 & 11.50   \\
                             & MonoScene + Ours & 50.44 & 17.08 \\ \cline{2-4}
                             & OccDepth~\cite{miao2023occdepth} & 41.60 & 12.84 \\
                             & OccDepth + Ours  & 50.20 & 16.79 \\ \hline
\multirow{4}{*}{Point Cloud} & SSA-SC~\cite{yang2021semantic}  & 58.25 & 24.54 \\
                             & SSA-SC + Ours   & 60.71 & 25.58 \\ \cline{2-4} 
                             & SCPNet~\cite{xia2023scpnet} & 50.24 & 37.55 \\
                             & SCPNet + Ours    & 59.25 & 38.19 \\ \hline
\end{tabular}
\caption{\textbf{Quantitative results of refining SSC on SemanticKITTI validation set~\cite{semantickitti}.} 
The results are based on the weights released by the authors on GitHub.}\label{tab:refine}
\end{center}
}
\end{table}

\paragraph{Scene Outpainting.}
Fig.~\ref{fig:city} illustrates a generated outpainting city-level scene, extending a $256\times256\times32$ scene to a substantial $1792\times3328\times32$ landscape.
Although our model was not designed to generate cityscapes, it demonstrates the capability of maintaining coherence over a large area.
The roads are connected in a meaningful and varied way, and various objects such as buildings, cars, and people have been created around them.
Please check the details of the scene in Fig.~\ref{fig:city}.
\begin{figure}[t]
 \centering
    \centering
    \captionsetup{type=figure}
        \includegraphics[trim={0cm 7.6cm 21.2cm 0cm},clip,width=1.0\columnwidth]{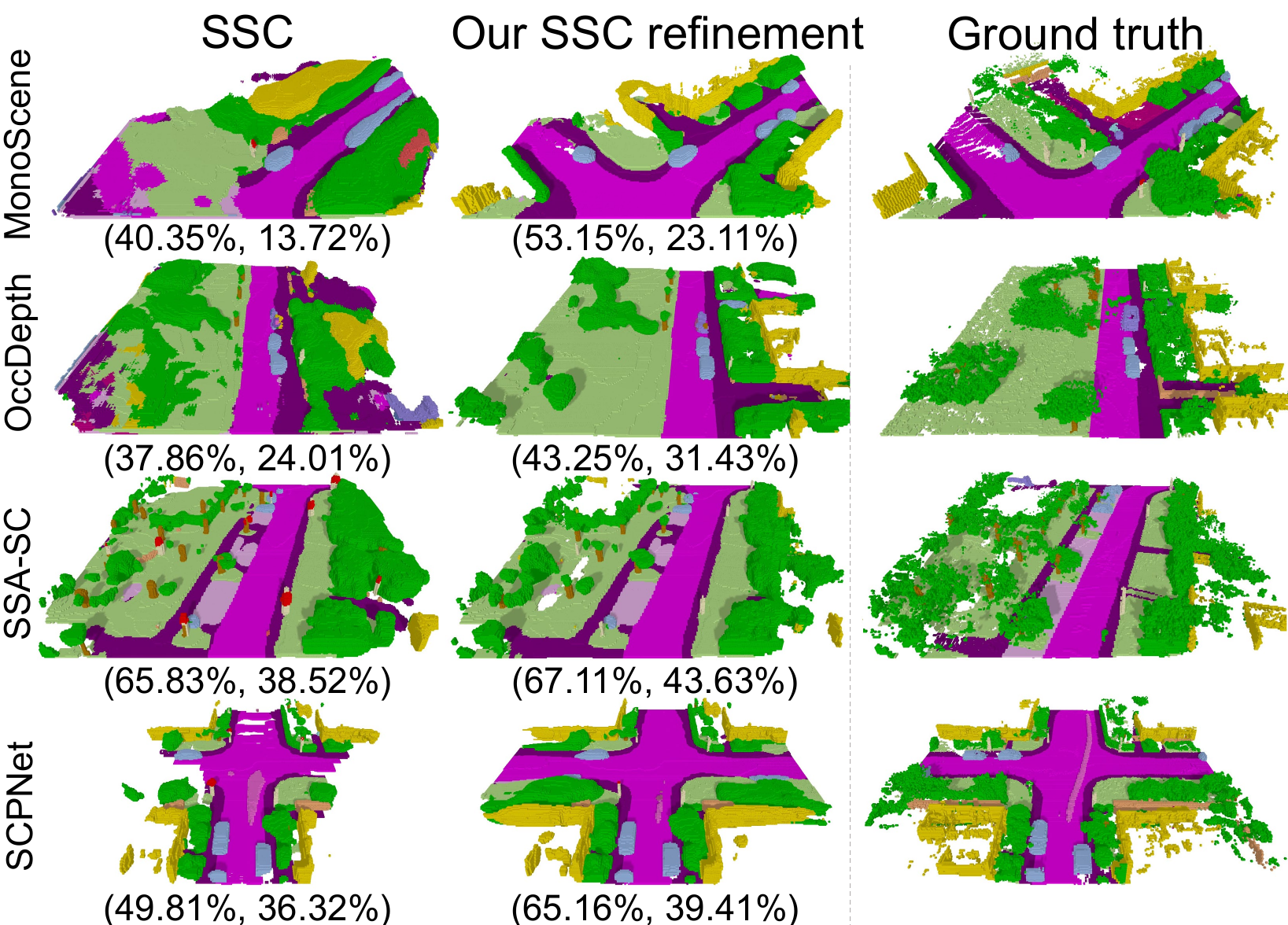}
        \caption{
            \textbf{Semantic scene completion refinement.}
            The SSC metrics are reported in the parentheses as (IoU, mIoU).
        }\label{fig:refine}  
\end{figure}

\paragraph{Semantic Scene Completion Refinement.}
In Fig.~\ref{fig:refine}, there is a notable semantic and geometric discrepancy between scenes predicted by existing semantic scene completion (SSC) methods and their real scene counterparts.
Our model helps to bridge this gap by employing a 3D scene prior, which is effectively modeled through our diffusion model.
SCPNet~\cite{xia2023scpnet} also attempts to learn geometric priors from a teacher model trained with merged sequential frames. 
Yet, its completion network is inclined toward conservative estimations, which frequently result in partially filled spaces.
In the case of SSA-SC~\cite{yang2021semantic}, which utilizes bird's-eye view features, it is sometimes inappropriate due to the label's inherent limitations of the bird's-eye perspective in capturing certain semantic details.
This discrepancy issue is even more evident in RGB-based methods, such as MonoScene~\cite{cao2022monoscene} and OccDepth~\cite{miao2023occdepth}, 
which often exhibit diminished sharpness when image features are projected into 3D space.
While SSC models show variations from real-world data distributions, our model shows potential in aligning these more closely with reality, as shown in Fig.~\ref{fig:refine}.

\begin{figure}[t]
 \centering
    \centering
    \captionsetup{type=figure}
        \includegraphics[trim={0cm 16.83cm 18.45cm 0cm},clip,width=1.0\columnwidth]{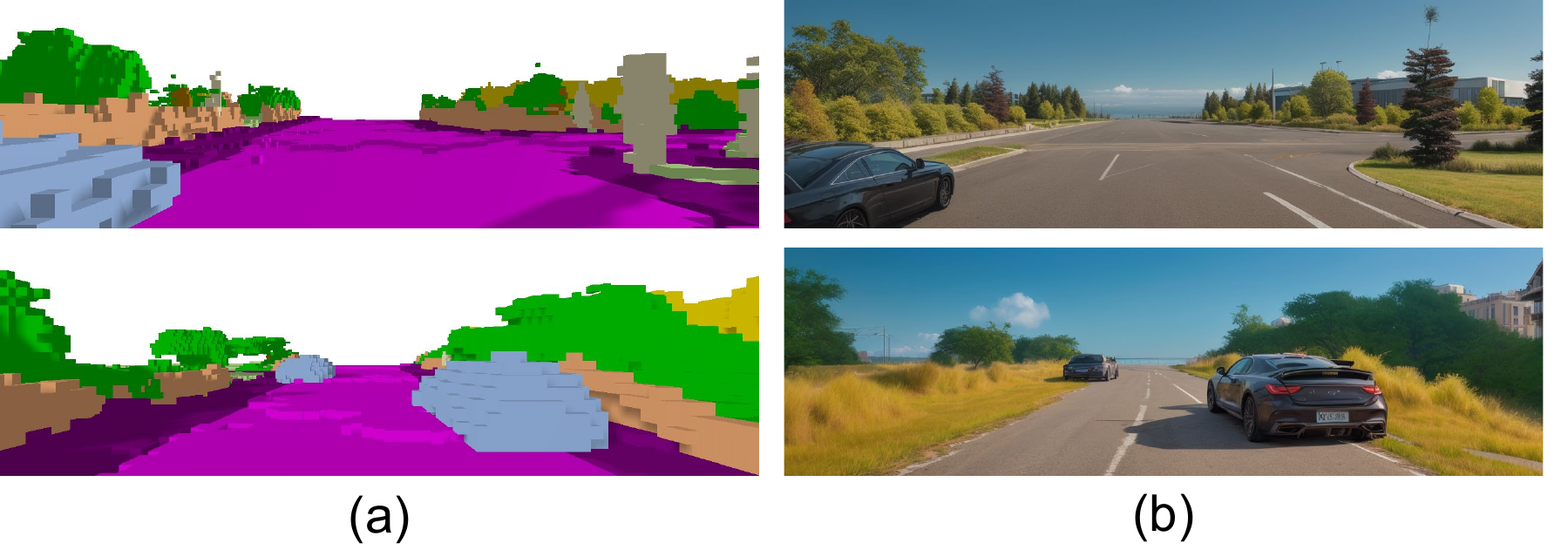}\vspace{-2mm}
        \caption{
            \textbf{Semantic scene to RGB.} 
            (a) Semantic maps are rendered into a driving view from our generated scenes.
            (b) We generate images through ControlNet~\cite{zhang2023adding}, an easily accessible image-to-image model, with generated semantic maps without shadows.
        }\label{fig:rgb}  
\end{figure}
As indicated in Tab.~\ref{tab:refine}, our SSC refinement process appears to offer improvements to all state-of-the-art SSC models. 
These preliminary results suggest our model's effectiveness in providing not just more accurate semantic segmentation but also a more complete scene.

\paragraph{Semantic Scene to RGB Image.}
As illustrated in Fig.~\ref{fig:rgb}, we conducted an image-to-image generation experiment.
The semantic maps are rendered into a driving view without shadows; then, they are utilized as inputs for ControlNet~\cite{zhang2023adding}. 
The generated RGB images are geometrically and semantically plausible but display a synthetic quality since the pretrained ControlNet was not trained on actual autonomous driving datasets.

\subsection{Ablation}

\begin{table}[]\renewcommand{\tabcolsep}{1.1mm}
\centering
\small{
\begin{tabular}{lccccc}
\hline
Method          & FID $\downarrow$   & KID $\downarrow$ & IS $\uparrow$   & Prec $\uparrow$ & Rec $\uparrow$  \\ \hline
Ours         & 56.55  & 0.04 & 3.25 & 0.39 & 0.32 \\ 
w/o $\mathtt{PE}$  & 59.14 & 0.04 & 3.21 & 0.30 & 0.31 \\ 
$xy$-plane (w/o triplane)   & 75.56  & 0.06 & 3.08 & 0.28 & 0.27 \\ 
3D-volume (w/o triplane)  & 136.50 & 0.15 & 1.96 & 0.14 & 0.12  \\ \hline
\end{tabular}
}
\caption{\textbf{Ablation studies on scene generation.} We ablate our model variants on SemanticKITTI~\cite{semantickitti}.}
\label{tab:ablation}\vspace{-2mm}
\end{table}



As shown in Tab.~\ref{tab:ablation}, we conducted ablation studies on our triplane diffusion model for scene generation, focusing on two key design elements:

\paragraph{Positional Embedding.} 
Our variant excluded positional embedding \texttt{PE}, which generates high-frequency features critical for detailed scene reconstruction. 
Its absence resulted in lower performance across all metrics.

\vspace{-3mm}

\paragraph{Triplane Representation.} 
We evaluated the effectiveness of the triplane representation for real-outdoor scene generation.
The triplane and $xy$-plane enhance the generation quality compared with 3D features, while the $xy$-plane shows lower performance than triplanes. 
We suspect that the excessive factorization limits the representation capability of the $xy$-plane compared to the triplane.

\subsection{Limitation}
While our model demonstrates significant progress in the generation of 3D real-outdoor scenes, it inherently reflects the characteristics of its training data, as detailed in Sec.~\ref{dataset}. 
This reliance introduces several limitations. 
One notable challenge is the model's difficulty in accurately depicting areas occluded from the sensor's viewpoint, such as the rear sides of buildings, often leading to their incomplete representation in our generated scenes. 
Moreover, since the dataset is captured from a driving view, there is an inherent shortfall in capturing the full height of buildings. 
This brings a partial representation of the vertical structure of buildings and other tall elements in the scenes.
Another issue is the model tends to produce traces of moving objects stemming from dataset pre-processing that merges sequential frames.
For future work, incorporating prior knowledge of the city into the model could yield better results, such as addressing occluded areas or building heights.

\section{Conclusion}
We have proposed a diffusion framework called \textit{SemCity} for real-world outdoor scene generation.
The seminal idea is to generate a scene by factorizing real-outdoor scenes into triplane representations.
Our triplane representation outperforms traditional voxel-based approaches, 
producing scenes that are not only visually more appealing but also rich in semantic detail, effectively capturing the complexity of various objects within the scene.
Ours is not constrained by fixed resolutions thanks to the incorporation of an implicit neural representation.
We have further expanded the capabilities of our triplane diffusion model to several practical applications, including scene inpainting, scene outpainting, and semantic scene completion refinement.
Specifically, by manipulating triplanes during the diffusion process, we achieve seamless inpainting and outpainting at both the object and scene levels. 
Ours is used to more closely align the scenes predicted by existing semantic scene completion methods with the actual data distribution using our learned 3D prior.
We believe that our work provides a road map of real-outdoor scene generation to research communities.


\twocolumn[{%
\renewcommand\twocolumn[1][]{#1}%
\begin{center}
   \Large \bf SemCity: Semantic Scene Generation with Triplane Diffusion\\- Supplementary Material -
\end{center}\vspace{2.5cm}
}]

\appendix
\setcounter{table}{0}
\setcounter{figure}{0}
\setcounter{equation}{0}
\makeatletter 
\renewcommand{\thefigure}{S\@arabic\c@figure}
\renewcommand{\thetable}{S\@arabic\c@table}
\renewcommand{\theequation}{S\@arabic\c@equation}
\makeatother

In this supplementary material, 
we report additional contents for an in-depth understanding of our method:
backgrounds for diffusion models (Sec.~\ref{supp:diffusion_background}),
implementation details of our method (Sec.~\ref{supp:implementation_details}),
and our additional experimental results (Sec.~\ref{supp:additional_exp_results}).
Specifically, we visualize our generation results across scene generation, scene inpainting, scene outpainting, and semantic scene completion refinement.
We further demonstrate RGB images generated from our scene samples.

\section{Backgrounds of Diffusion Models}\label{supp:diffusion_background}
Diffusion models synthesize data (\textit{e.g.}, images) by gradually transforming a random noise distribution into a data distribution through a reverse Markov process. 
This process involves two main phases: the forward process (\textit{i.e.}, diffusion process) and the reverse process (\textit{i.e.}, denoising process).

\subsection{Forward Process}
In the forward process, a given data $\mathbf{x}_0 \sim p(\mathbf{x}_{0})$ is gradually corrupted by adding noise over a series of steps. 
This process transforms the original data distribution into a Gaussian distribution.
The forward process is modeled as a Markov chain, where each step adds a small amount of noise, making it easy to compute and invert:
\begin{equation}\label{eq:forward_1step}
q(\mathbf{x}_{t}|\mathbf{x}_{t-1}) = \mathcal{N}(\sqrt{1 - \beta_{t}}\mathbf{x}_{t-1}, \beta_{t}\mathbf{I}).
\end{equation}
Here, $\mathbf{x}_{t}$ is a noised data at step ${t}$, $\beta_{t}$ is a variance schedule, and $\mathcal{N}$ denotes the Gaussian distribution.
$t$ is defined within $1 \le t \le T$ with the maximum denoising steps $T$.

The $t$-th noised data $\mathbf{x}_t$ is sampled via iteration of the forward process $q(\mathbf{x}_{t}|\mathbf{x}_{t-1})$ in Eq.~\ref{eq:forward_1step}; 
however, $\mathbf{x}_t$ can be simply obtained as a closed form with ${\alpha}_t = 1 - \beta_{t}$ and $\bar{\alpha}_t = \Pi^t_{s=0}\alpha_s$:
\begin{align}
q(\mathbf{x}_t \vert \mathbf{x}_0) &= \mathcal{N}(\sqrt{\bar{\alpha}_t} \mathbf{x}_0, (1 - \bar{\alpha}_t)\mathbf{I})  \label{eq:forward_multistep1}, \\
\mathbf{x}_t &= \sqrt{\bar{\alpha}_t} \mathbf{x}_0 + \bm{\epsilon} \sqrt{(1 - \bar{\alpha}_t)}, \label{eq:forward_multistep2}
\end{align}
where $\bm{\epsilon} \sim \mathcal{N}(\mathbf{0}, \mathbf{I})$, and $1 - \bar{\alpha}_t$ is a variance of the noise for an arbitrary timestep $t$.

\subsection{Reverse Process}
The reverse process iteratively removes noises from the sample
to generate a coherent structure resembling the original data $\mathbf{x}_{0}$ distribution. 
Each denoising step can be expressed as a reverse Markov chain:
\vspace{-2mm}
\begin{equation}
p_\phi(\mathbf{x}_{t-1}|\mathbf{x}_{t}) = \mathcal{N}(\bm{\mu}_{\phi}(\mathbf{x}_{t}, t), \bm{\Sigma}_{\phi}(\mathbf{x}_{t}, t)), 
\label{eq:reverse_diffusion}
\end{equation}
where $\bm{\mu}_{\phi}$ and $\bm{\Sigma}_{\phi}$ are the mean and covariance of the reverse process at step ${t}$, parameterized by learnable parameters ${\phi}$.
In particular, \cite{ho2020denoising} proposes that a model $\bm{\epsilon}_\phi(\mathbf{x}_t, t)$ can simply be trained to predict the noise $\bm{\epsilon}$ instead of directly parameterizing the mean $\bm{\mu}_{\phi}(\mathbf{x}_{t}, t)$.
They assume the covariance $\bm{\Sigma}_{\phi}(\mathbf{x}_{t}, t)$ is constant.
Thus, we can define a diffusion loss as:
\vspace{-2mm}
\begin{equation}
\label{eq:loss}
    \mathcal{L} = \mathbb{E}_{t \sim \mathcal{U}(1,T), \bm{\epsilon} \sim \mathcal{N}(\mathbf{0}, \mathbf{I})} || \bm{\epsilon} - \bm{\epsilon}_{\phi}(\mathbf{x}_{t}, t)||_{2},
\end{equation}
where $\mathcal{U}$ is the discrete uniform distribution.
\cite{austin2021structured} suggests the $\mathbf{x}_0$-parameterization where a model $\mathbf{x}_{\phi}$ predicts the input data $\mathbf{x}_{0}$ directly, rather than predicting the added noise $\bm{\epsilon}$.
The diffusion loss for the $\mathbf{x}_0$-parameterization is defined as:
\begin{equation}
\label{eq:x0_loss}
    \mathcal{L} = \mathbb{E}_{t \sim \mathcal{U}(1,T)}||\mathbf{x}_0 - \mathbf{x}_{\phi}(\mathbf{x}_{t}, t)||_{2}.
\end{equation}
This loss function is the basis of our triplane diffusion loss in Eq.~2 of the main paper.
\section{Implementation Details}\label{supp:implementation_details}
\subsection{Training Setting}
\paragraph{Triplane Autoencoder.}
As described in Sec.~3.1 of the main paper, our triplane autoencoder consists of two modules: the triplane encoder $f_\theta$ and the implicit MLP decoder $g_\theta$.
We configure the encoder $f_\theta$ with six 3D convolutional layers with a skip connection
and design our MLP decoder $g_\theta$ to be light to mitigate the training burden. The MLP decoder consists of four 128-dimensional fully-connected layers with a skip connection.
Following ~\cite{mildenhall2020nerf}, the positional encoding $\texttt{PE}(\mathbf{p})$ at coordinates $\mathbf{p}$ is used as sinusoidal functions defined as: $\texttt{PE}(\mathbf{p}) = \left[ \sin(2^0\pi \mathbf{p}), \cos(2^0\pi \mathbf{p}), \ldots, \sin(2^{5}\pi \mathbf{p}), \cos(2^{5}\pi \mathbf{p}) \right]$.

\paragraph{Triplane Diffusion Model.}
Based on the observation~\cite{saharia2022palette} where the sample diversity depends on $L_1$ or $L_2$ diffusion loss, 
the norm factor $p$ of the triplane diffusion loss (Eq.~2 of the main paper) is set to $1$ or $2$.
For more diversity of generation results, we set $p=2$ (\textit{i.e.}, $L_2$) in scene generation, scene inpainting, and scene outpainting.
In contrast, we use $p=1$ (\textit{i.e.}, $L_1$) for semantic scene completion refinement following \cite{saharia2022image}.
The diffusion settings (\textit{e.g.}, the variance schedule $\beta_t$) are used as DDPM~\cite{ho2020denoising}.

\subsection{Generation Setting}

\paragraph{Scene Outpainting.}
Our model extrapolates a given scene, resulting in a larger scale scene as depicted in Fig.~\ref{sup_fig:outpainting}, Fig.~\ref{sup_fig:city} and Fig.~6 of the main paper. 
As shown in Fig.~\ref{sup_fig:outpainting}, our model is capable of generating a variety of extended scenes.  
To enhance its effectiveness, we have incorporated an interactive outpainting system~\cite{Outpainting} that allows users to guide the scene generation process. 
This interaction is a demonstration of the model's flexibility and responsiveness to user preferences. 
Users may keep the original outpainting or regenerate it to correspond more closely to their visual objectives. 
This capability enables users to create finely-tuned urban scenes on a city-scale, as shown in Fig.~\ref{sup_fig:city} and Fig.~6 of the main paper. 

\paragraph{Semantic Scene to RGB Image.}
We exploit ControlNet~\cite{zhang2023adding}
to generate RGB images from our semantic scenes.
ControlNet supports various conditional inputs (\textit{e.g.}, segmentation or depth maps) and can be easily integrated with other fine-tuned models (\textit{e.g.}, Dreambooth~\cite{ruiz2023dreambooth}, Textual inversion~\cite{gal2022textual}, and Lora~\cite{hu2022lora}).
We manipulate a semantic map rendered from our generated scenes and generate an RGB image through the following process.
An initial RGB image is obtained by conditioning semantic and depth maps rendered from our generated scene.
Afterward, we generate a final image from the initial RGB map with conditional segmentation and depth maps obtained from ControlNet preprocessors~\cite{zhou2017scene, zhou2019semantic, Ranftl2022}.
For our experiments, we employ the diffusion model~\cite{Rombach_2022_CVPR} weights\footnote{https://civitai.com/models/119169/urban-streetview} fine-tuned on urban street views to generate images analogous to driving scenes.

\section{Additional Experimental Results}\label{supp:additional_exp_results}

In this section, we visualize additional generated scenes of our method in the various applications, including 1) scene generation, 2) scene inpainting, 3) scene outpainting, 4) semantic scene completion refinement, and 5) semantic scene to RGB image.
For visualizations, colors are used as below.
\begin{figure}[!h]
\vspace{-4mm}
\begin{center}
\centerline{\includegraphics[trim={0cm 56.18cm 30.08cm 0cm},clip,width=0.9\columnwidth]{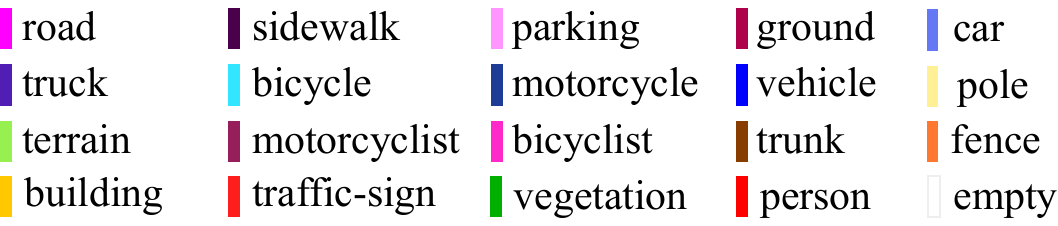}}
\end{center}
\vspace{-10mm}
\end{figure}

\newpage
\subsection{Triplane Visualization}
\vspace{-3mm}
\begin{figure}[!h]
\begin{center}
\centerline{\includegraphics[trim={0cm 6.45cm 27.4cm 0cm},clip,width=0.83\columnwidth]{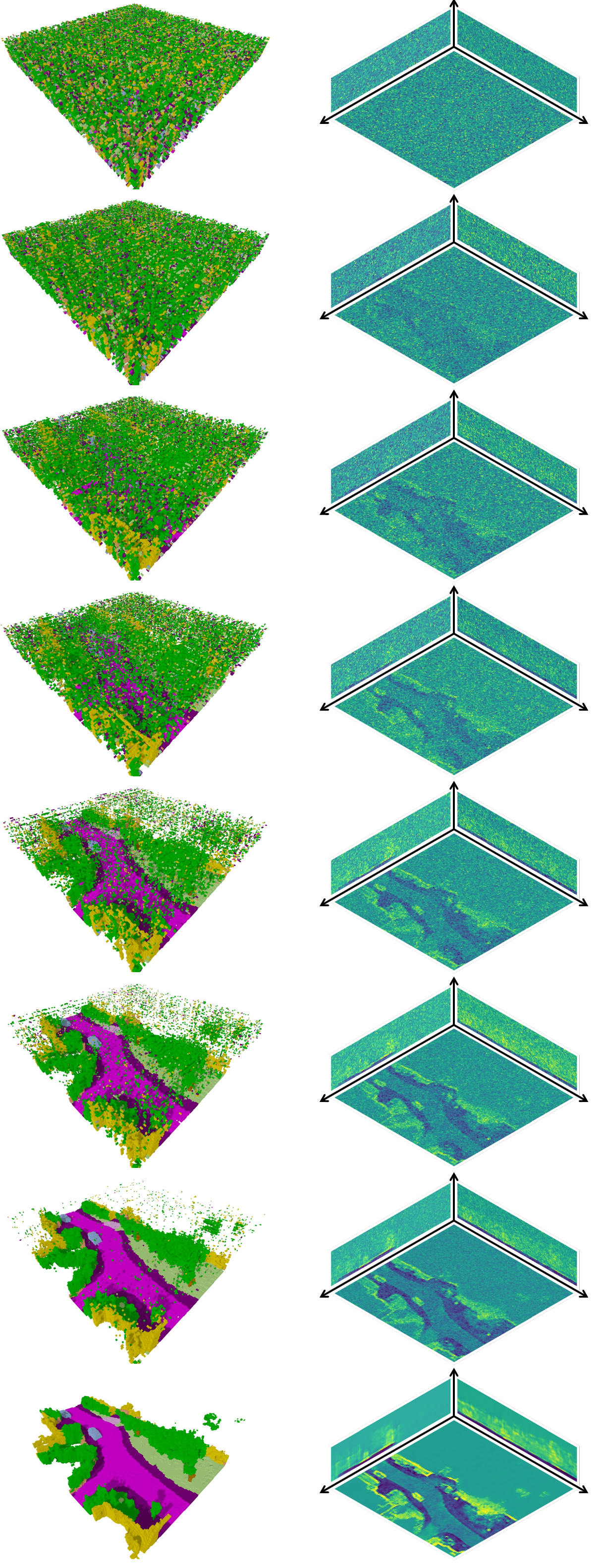}}
    \captionof{figure}{
    \textbf{Triplane visualization during our generation process.}
    We visualize triplanes (right) and their corresponding scenes (left) according to diffusion steps.
    We observe distinct denoising patterns where our diffusion model initially constructs low-frequency structures (\textit{e.g.}, roads) in the early stages of denoising. In contrast, high-frequency details (\textit{e.g.}, edges) are progressively refined in the later stages of the process. This phenomenon can also be found in image diffusion models~\cite{ho2020denoising}; we expect this property to be
    exploited for elastic scene editing in future work.
    }\label{supp_fig:triplane}
\end{center}
\vspace{-10mm}
\end{figure}

\clearpage
\noindent\begin{minipage}{\textwidth}
    \subsection{Scene Generation}

\begin{center}
\centerline{\includegraphics[trim={0cm 0cm 0.0cm 0cm},clip,width=0.95\textwidth]{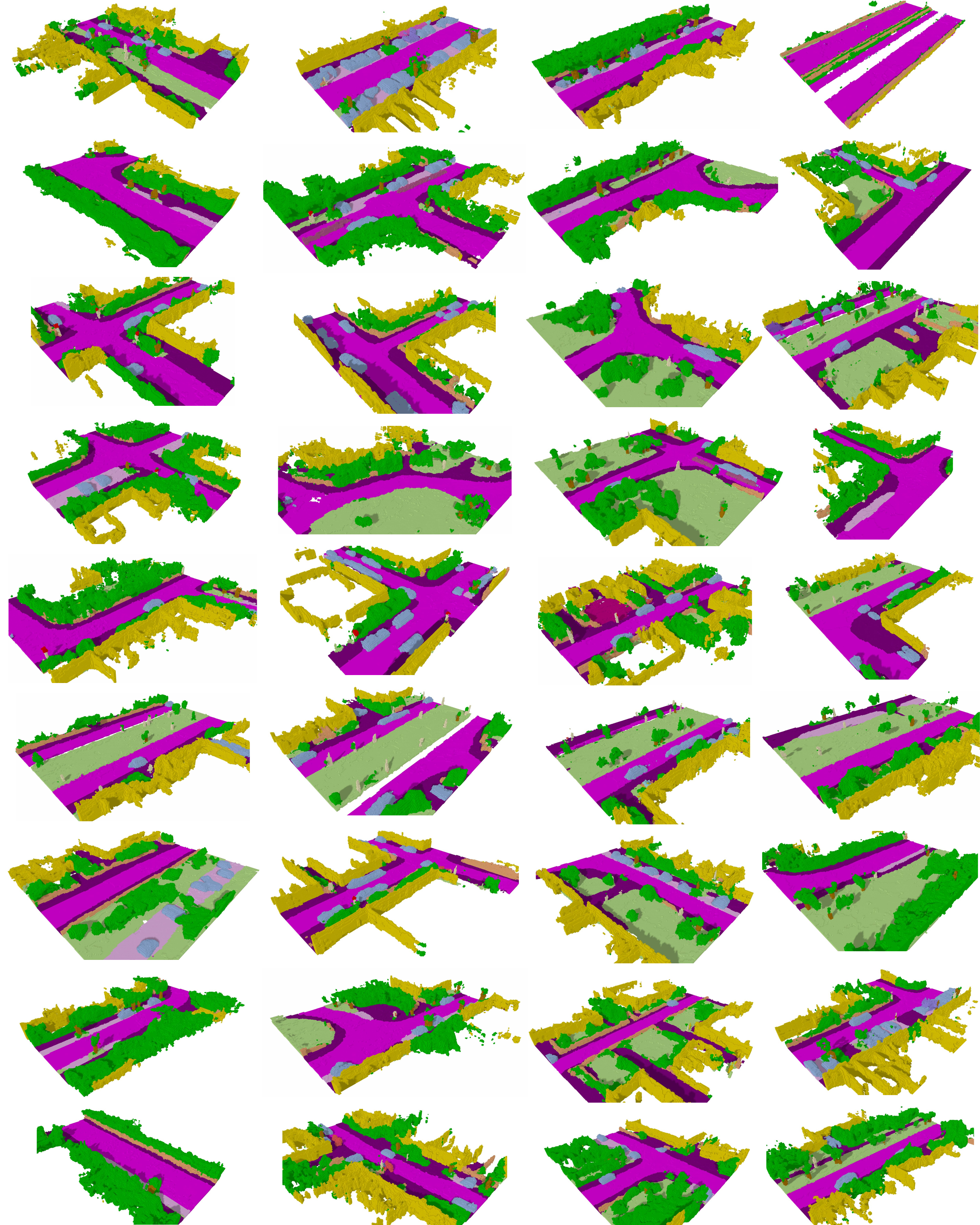}}
    \captionof{figure}{
    \textbf{Scene generation results of our method.}
    The generated scenes demonstrate various road shapes, including L, T, Y, straight, and crossroads, which show that our method generates diverse samples.
    }
\end{center}

\end{minipage}

\clearpage
\noindent\begin{minipage}{\textwidth}
    \subsection{Semantic Scene Completion Refinement}

\begin{center}
\centerline{\includegraphics[trim={0cm 0cm 0.0cm 0cm},clip,width=0.93\textwidth]{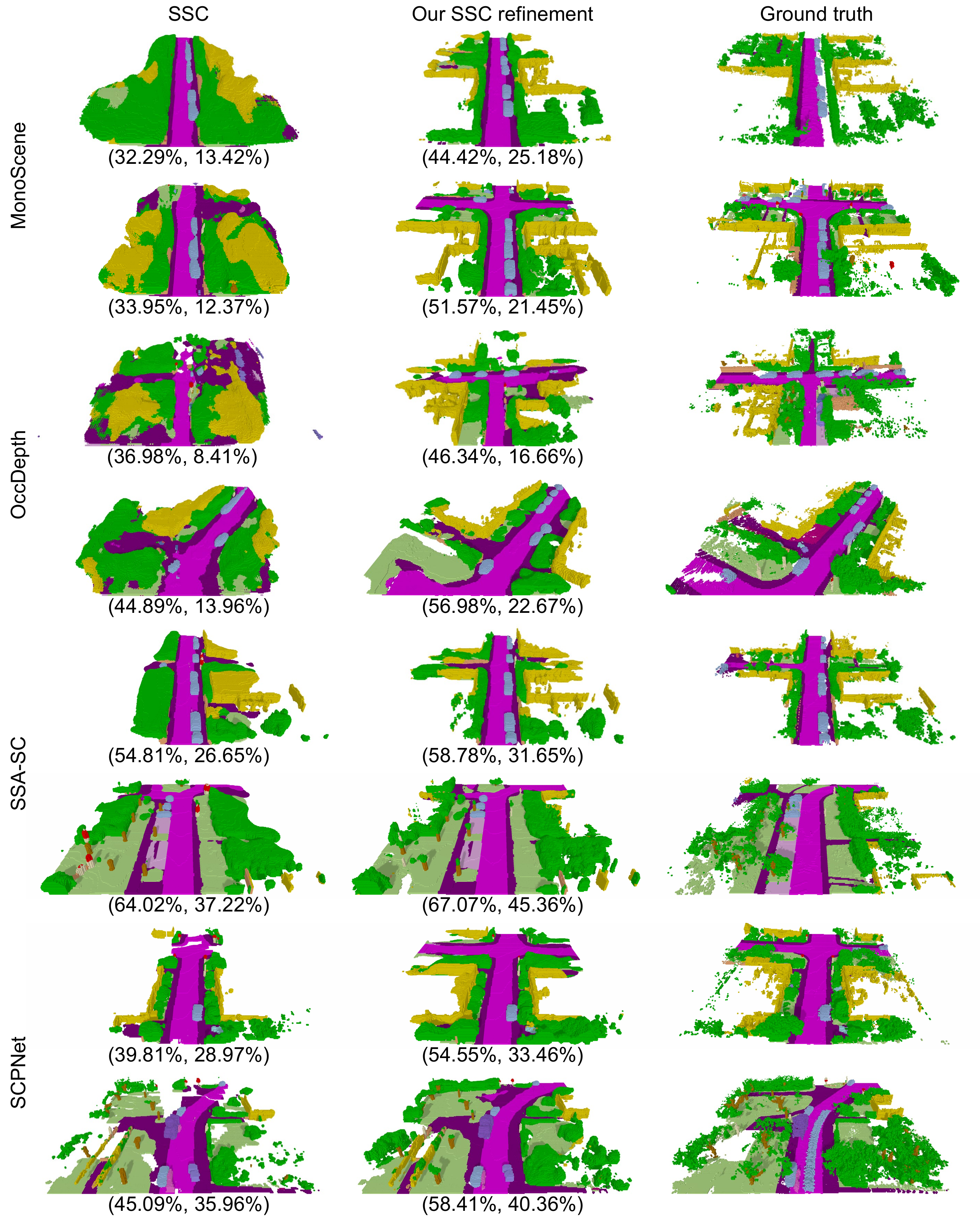}}
    \captionof{figure}{
    \textbf{Results of semantic scene completion refinement of our method.}
    The parentheses report the SSC metrics as (IoU, mIoU).
    Our method refines the results of state-of-the-art SSC methods.
    The MonoScene~\cite{cao2022monoscene} and OccDepth~\cite{miao2023occdepth} methods use a RGB input. The SSA-SC~\cite{yang2021semantic} and SCPNet~\cite{xia2023scpnet} employ LiDAR point clouds as an input.
    }
\end{center}

\end{minipage}

\clearpage
\noindent\begin{minipage}{\textwidth}
    \subsection{Scene Outpainting}
    \begin{center}
\centerline{\includegraphics[trim={0cm 0cm 0.0cm 0cm},clip,width=0.95\textwidth]{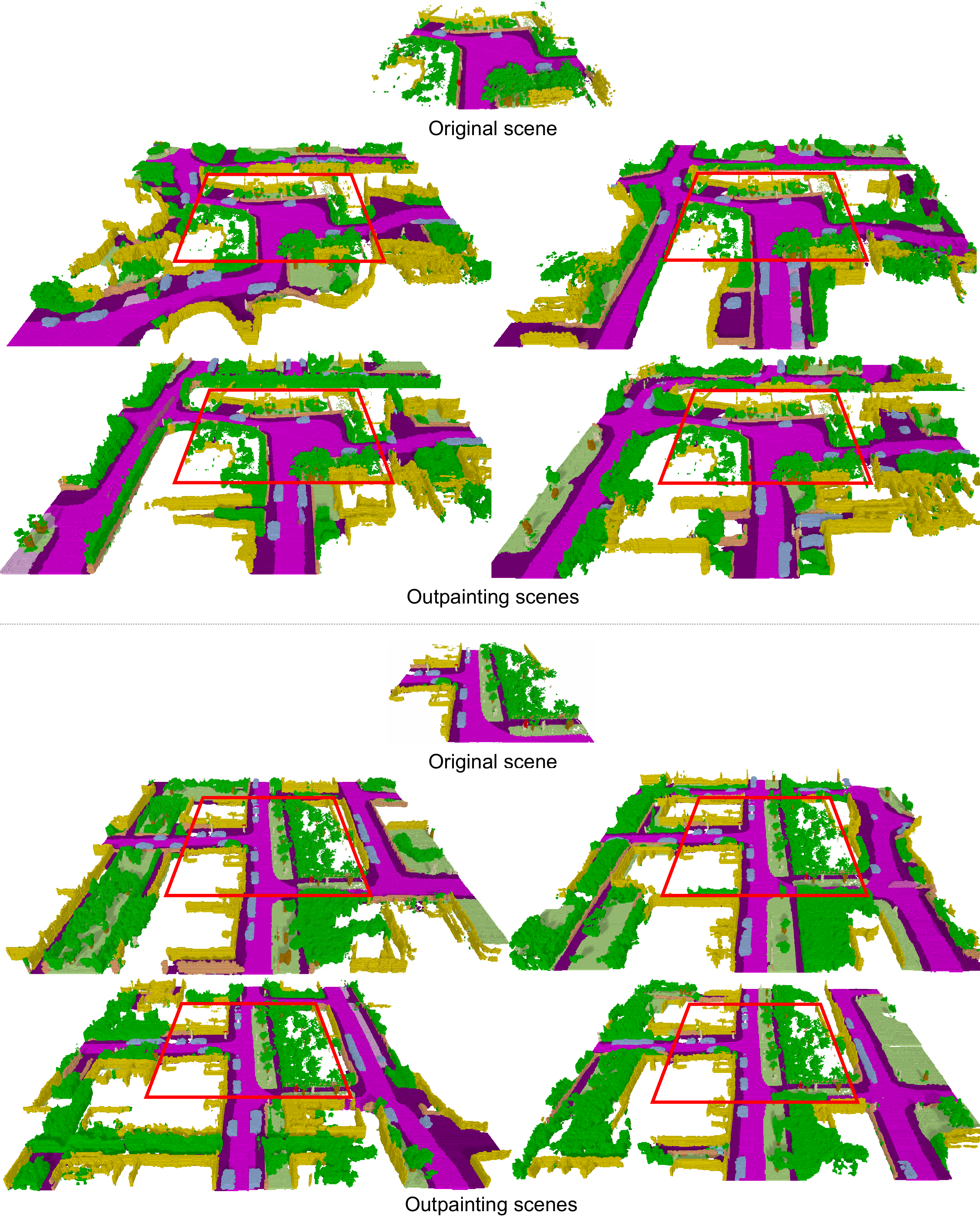}}
    \captionof{figure}{
    \textbf{Scene outpainting results of our method.}
    We visualize various outpainting results generated from two scenes. The outpainted scene is expanded from the given size of $256 \times 256 \times 32$ to $512 \times 512 \times 32$ without any guidance. 
    The red boxes mean an original scene for outpainting.
    Our method produces various outpainted scenes from an identical original scene.
    }
    \label{sup_fig:outpainting}
    \vspace{-1cm}
\end{center}

\end{minipage}

\clearpage
\noindent\begin{minipage}{\textwidth}
    \subsection{City-level Generation}
    \begin{center}
\centerline{\includegraphics[trim={0cm 0.1cm 0.0cm 0cm},clip,width=0.95\textwidth]{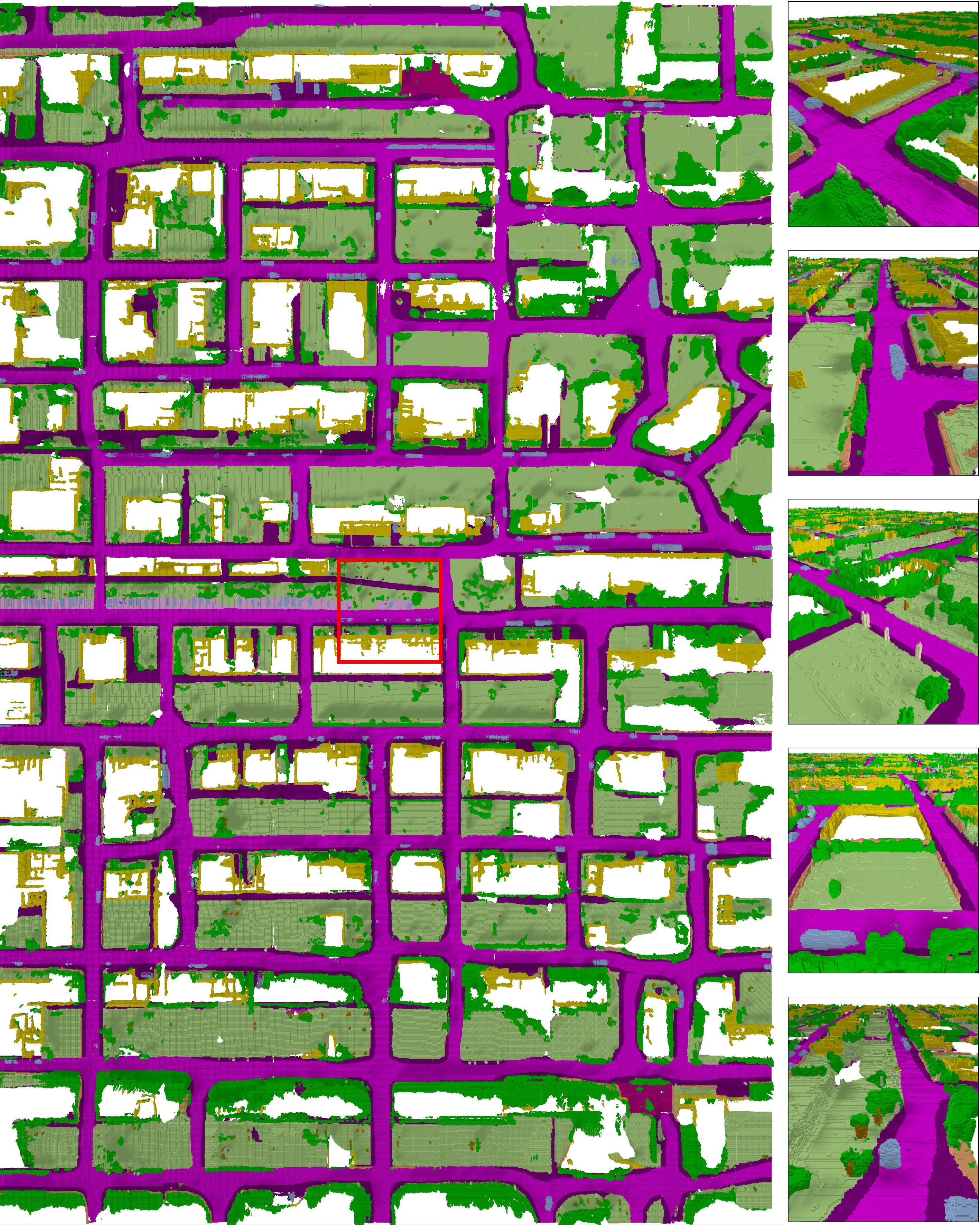}}
    \captionof{figure}{
    \textbf{City-scale outpainted scene.}
    The first column displays a city-scale scene, showcasing an expansive urban landscape.     
    The city-scale scene is expanded from the original size of $256 \times 256 \times 32$ to $1792 \times 2816 \times 32$.
    The second column figures provide close-up views of specific areas within the city-scale scene.
    The red box means an original scene for outpainting.
    }
    \label{sup_fig:city}
    \vspace{-10mm}
\end{center}

\end{minipage}

\clearpage
\noindent\begin{minipage}{\textwidth}
    \subsection{Scene Inpainting}

\begin{center}
\centerline{\includegraphics[trim={0cm 20.77cm 0.0cm 0cm},clip,width=1.0\textwidth]{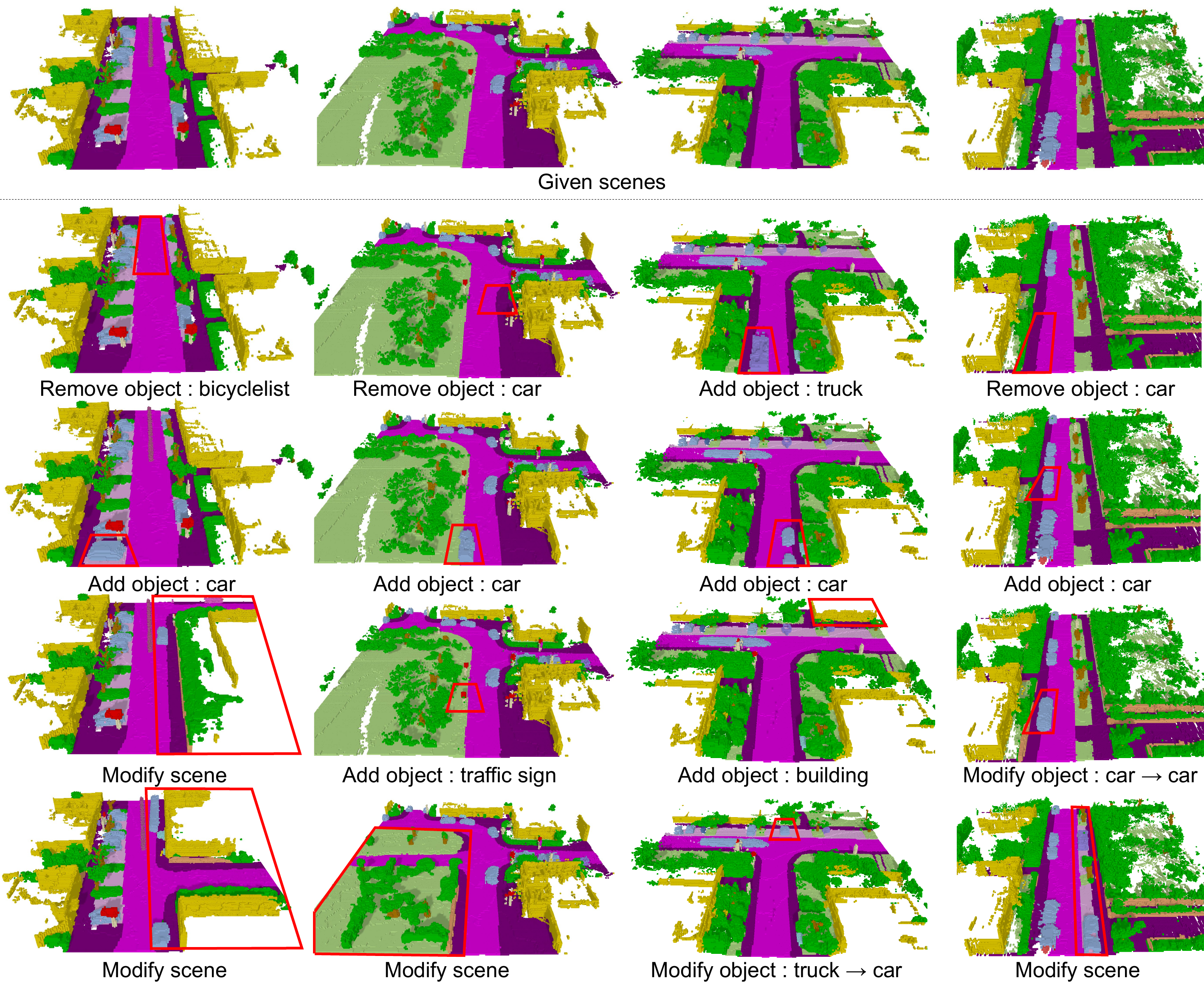}}
    \captionof{figure}{
    \textbf{Scene inpainting results of our method.}
    The red boxes refer to inpainting regions.
    }
    \label{supp_fig:inpaint}
\end{center}

    \vspace{-3mm}
    \subsection{Semantic Scene to RGB Image}
    \begin{center}
\centerline{\includegraphics[trim={0cm 47.6cm 0.0cm 0cm},clip,width=1.0\textwidth]{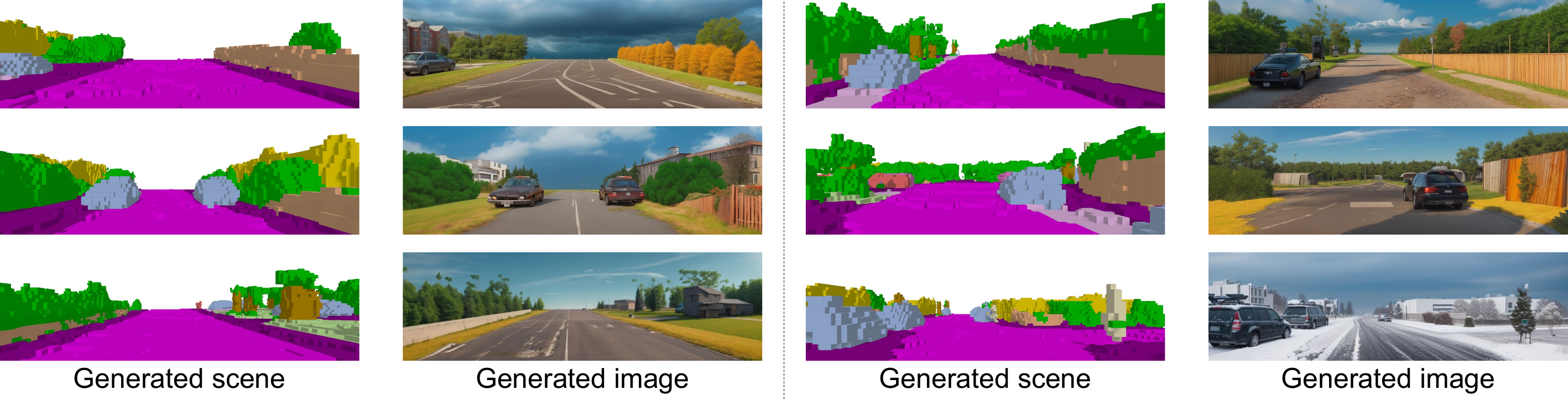}}
    \captionof{figure}{
    \textbf{RGB images generated from our generated scenes.}
    ControlNet~\cite{zhang2023adding} is utilized to generate images from our generated scenes.
    In the last figure illustrating a snowy scene, we added a text prompt `snow'.
    }
\end{center}
    \vspace{-10mm}
\end{minipage}

\clearpage

{
\noindent \textbf{\\Acknowledgement.}
This work was supported by IITP (Institute of Information \& Communications Technology Planning \& Evaluation) and ITRC (Information Technology Research Center), funded by Korea government (MSIT) (RS-2023-00237965(2024) and IITP-2024-2020-0-01460).
Prof. Sung-Eui Yoon is a corresponding author. 
}

\thispagestyle{empty}

{
    \small
    \bibliographystyle{ieeenat_fullname}
    \bibliography{main}
}

\end{document}